\documentclass[letterpaper]{article} 
\usepackage{aaai24}  
\usepackage{times}  
\usepackage{helvet}  
\usepackage{courier}  
\usepackage[hyphens]{url}  
\usepackage{graphicx} 
\usepackage{natbib}  
\usepackage{caption} 
\usepackage{algorithm}
\usepackage{algorithmic}
\usepackage{url}            
\usepackage{booktabs}       
\usepackage{multirow}
\usepackage{amsmath}
\usepackage{amsfonts}       
\usepackage{cleveref}
\usepackage{subcaption}
\usepackage[utf8]{inputenc}

\usepackage{newfloat}
\usepackage{listings}
\DeclareCaptionStyle{ruled}{labelfont=normalfont,labelsep=colon,strut=off} 
\floatstyle{ruled}
\newfloat{listing}{tb}{lst}{}
\floatname{listing}{Listing}
\title{FedTabDiff: Federated Learning of Diffusion Probabilistic Models \\ for Synthetic Mixed-Type Tabular Data Generation}
\author{
    \hspace{2em} Timur Sattarov\textsuperscript{\rm 1, \rm 2 \rm*} \hspace{6em}
    Marco Schreyer\textsuperscript{\rm 1, \rm 3 \rm*} \hspace{5em}
    Damian Borth\textsuperscript{\rm 1} \hspace{1em} \\
    {\small \texttt{timur.sattarov@bundesbank.de \hspace{2em} marco@icsi.berkeley.edu \hspace{1em} damian.borth@unisg.ch}} \\
}
\affiliations{
    \textsuperscript{\rm 1}School of Computer Science, University of St.Gallen, St.Gallen, Switzerland

    \textsuperscript{\rm 2}Research Data and Service Centre, Deutsche Bundesbank, Frankfurt am Main, Germany\\

    \textsuperscript{\rm 3}International Computer Science Institute (ICSI), Berkeley, USA\\
    
}

\begin{document}

\maketitle

\begin{abstract}
Realistic synthetic tabular data generation encounters significant challenges in preserving privacy, especially when dealing with sensitive information in domains like finance and healthcare. In this paper, we introduce \textit{Federated Tabular Diffusion} (FedTabDiff) for generating high-fidelity mixed-type tabular data without centralized access to the original tabular datasets. Leveraging the strengths of \textit{Denoising Diffusion Probabilistic Models} (DDPMs), our approach addresses the inherent complexities in tabular data, such as mixed attribute types and implicit relationships. More critically, FedTabDiff realizes a decentralized learning scheme that permits multiple entities to collaboratively train a generative model while respecting data privacy and locality. We extend DDPMs into the federated setting for tabular data generation, which includes a synchronous update scheme and weighted averaging for effective model aggregation. Experimental evaluations on real-world financial and medical datasets attest to the framework's capability to produce synthetic data that maintains high fidelity, utility, privacy, and coverage.
\end{abstract}

\section{Introduction}

In the rapidly changing financial regulatory landscape, data analytics has become increasingly vital, with central banks collecting extensive microdata to guide policy, assess risks, and maintain stability globally. The detailed nature of this data introduces unique challenges, notably in data privacy. 
Real-world tabular data is crucial for developing complex, real-life dynamic models but often contains sensitive information, necessitating strict regulations like the EU's \textit{General Data Protection Regulation} and the \textit{California Privacy Rights Act}. Despite implementing such safeguards, institutions remain wary of deploying \textit{Artificial Intelligence}~(AI) models due to risks of data leakage~\cite{carlini2020, bender2021, kairouz2019} and model attacks~\cite{fredrikson2015, shokri2017, salem2018}, which could potentially reveal personally identifiable information or confidential training data~\cite{carlini2020, bender2021, kairouz2019, fredrikson2015, shokri2017, salem2018}. 

\vspace{0.1cm}

A promising avenue to address these risks originates from generating high-quality synthetic data. In this context, the term \textit{`synthetic data'} denotes data of a generative process grounded in the inherent properties of real data. Modeling these processes provides a nuanced understanding of underlying patterns, unlocking insights, especially in high-stake domains~\cite{assefa2020generating, schreyer2019b}.  This idea stands apart from conventional data obfuscation methods such as anonymization or eliminating sensitive attributes. Within high-stake domains, the generation of high-fidelity synthetic tabular data finds its rationale in:

\begin{itemize}

\item \textbf{Data Sharing:} Synthetic data enables data sharing that complies with regulatory mandates and unlocks the collaboration between the scientific community, domain experts, or other institutions.

\item \textbf{Data Liberation:} Synthetic data alleviates data usage restrictions, fostering a flexible environment for data analysis without breaching confidentiality agreements or regulatory boundaries.

\end{itemize}

\noindent Generating high-fidelity synthetic tabular data is central for regulatory-compliant data sharing, fostering collaboration among various entities, and enabling the modeling of rare but impactful events like fraud~\cite{charitou2021synthetic, Barse2003} or diseases. It is particularly pertinent as real-world tabular data is often characterized by inherent complexities:


\begin{enumerate}
    \item \textbf{Mixed Attribute Types:} Tabular data comprises diverse attribute data types, including categorical, numerical, and ordinal data distributions. Modeling these complexities requires effectively integrating different data types into a cohesive generative model.
    \item \textbf{Implicit Relationships:} Tabular data encompasses implicit relationships between individual records and attributes. Modeling these complexities necessitates disentangling the dependencies into a model representing the relationships between records and attributes. 
    \item \textbf{Distribution Imbalance:} The skewed distributions and imbalances in real-world tabular data represent a significant challenge. These factors demand advanced modeling techniques that precisely encapsulate and represent the nuanced patterns in tabular data.
\end{enumerate}


\begin{figure*}[t!]
  \centering
  \includegraphics[width=0.9\linewidth, keepaspectratio]{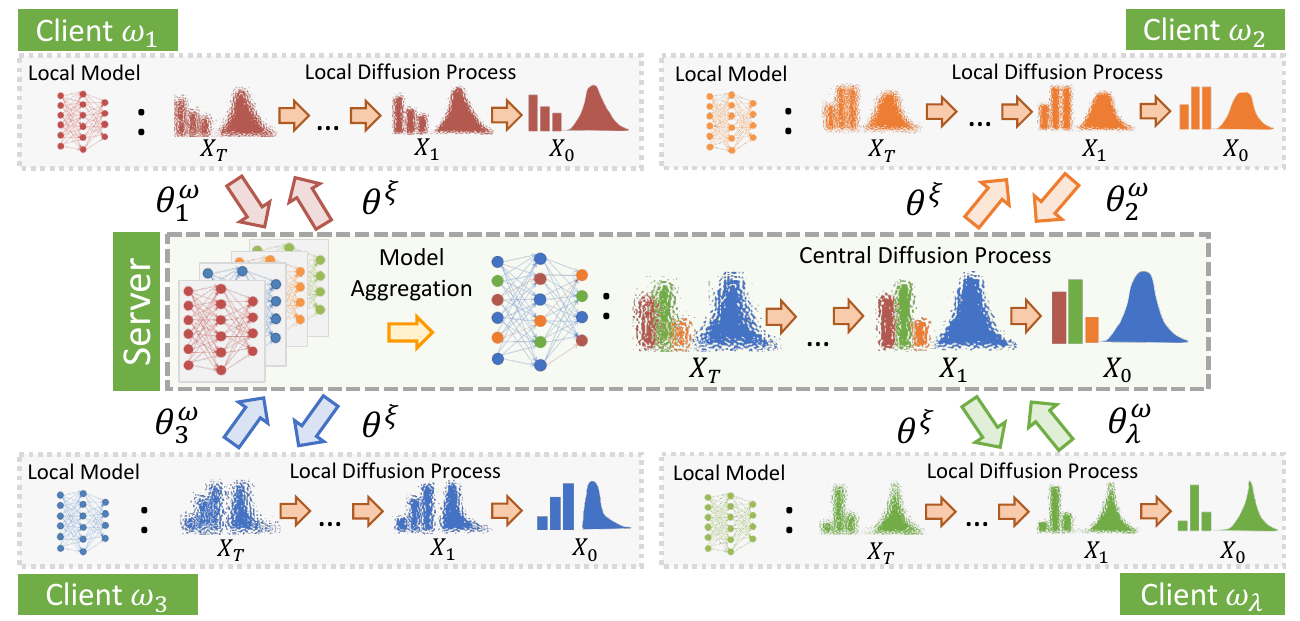}
    \caption{Schematic representation of the proposed \textit{FedTabDiff} model. It illustrates how each client $\omega_{i}$ independently trains a local diffusion model, named \textit{FinDiff}~\cite{sattarov2023findiff}. The process is depicted through various timesteps $X_T, \ldots, X_1, X_0$, each representing different stages of latent data representations in the reverse diffusion process. The individual model parameters, denoted as $\theta_{i}^{\omega}$, are periodically aggregated on a central server to form the consolidated model $\theta^{\xi}$. After every training round, the server then redistributes this consolidated model to each client.}
  \label{fig:FedTabDiff}
\end{figure*}

\noindent In recent years, deep generative models~\cite{goodfellow2014} have made impressive strides in the creation of diverse and realistic content, such as images~\cite{brock2018large, rombach2022high}, videos~\cite{chan2019everybody, singer2022make, yu2023magvit}, audio~\cite{wang2023neural, le2023voicebox, bai20223}, code~\cite{li2022competition, chen2022codet, le2022coderl}, and natural language~\cite{openai2023gpt4, bubeck2023sparks, touvron2023llama}. Lately, \textbf{\textit{Denoising Diffusion Probabilistic Models~(DDPM)}}~\cite{sohl2015deep, ho2020denoising} demonstrated the ability to generate synthetic images of exceptional quality and realism~\cite{dhariwal2021diffusion, rombach2022high, gao2023masked}. To effectively train these models, characterized by extensive parameters, substantial compute resources, and extensive training data are crucial~\cite{de2023training}. However, challenges arise when sensitive data is distributed across various institutions, such as hospitals, municipalities, and financial authorities, and cannot be shared due to privacy concerns~\cite{assefa2020generating, schreyer2022federatedcontinual, schreyer2022federated}. Recently, the concept of \textbf{\textit{Federated Learning}~(FL)}~\cite{mcmahan2017a, mcmahan2017b} was proposed in which multiple devices, such as smartphones or servers, collaboratively train AI models under the orchestration of a central server~\cite{kairouz2019}. The training data remains decentralized, providing a pathway to learn a shared model without direct data exchange.

\vspace{0.1cm}

This study addresses the challenge of learning models for mixed-type tabular data generation while maintaining data privacy. The main contributions are:

\begin{itemize}
    \item Introduction of \textit{FedTabDiff}, a novel federated learning framework for generating synthetic mixed-type tabular data, merging DDPMs and FL to enhance data privacy.
    \item Empirical evaluation of \textit{FedTabDiff} using real-world financial and healthcare datasets, illustrating its efficacy in synthesizing high-quality, privacy-compliant data.
\end{itemize}

\noindent The reference implementation of the code can be accessed through: \url{https://github.com/sattarov/FedTabDiff}.


\section{Related Work}

Lately, diffusion models\cite{cao2022survey, yang2022diffusion, croitoru2023diffusion} and federated learning~\cite{aledhari2020federated, li2021survey, zhang2021survey} triggered a considerable amount of research. In the realm of this work, the following review of related literature focuses on the federated deep generative modeling of tabular data: 

\vspace{0.1cm}

\textbf{Deep Generative Models:} Xu et al.~\cite{tvae_ctgan} introduced CTGAN, a conditional generator for tabular data, addressing mixed data types to surpass previous models' limitations. Building on GANs for oversampling, Engelmann and Lessmann~\cite{engelmann2021conditional} proposed a solution for class imbalances by integrating conditional Wasserstein GANs with auxiliary classifier loss. Jordon et al.~\cite{jordon2018pate} formulated PATE-GAN to enhance data synthesis privacy, providing differential privacy guarantees by modifying the PATE framework. Torfi et al.~\cite{torfi2022differentially} presented a differentially private framework focusing on preserving synthetic healthcare data characteristics. To handle diverse data types more efficiently, Zhao et al.~\cite{zhao2021ctab} developed CTAB-GAN, a conditional table GAN that efficiently addresses data imbalance and distributions. Kim et al.~\cite{kim2021oct} enhanced synthetic tabular data utility using neural ODEs, and Wen et al.~\cite{wen2022causal} introduced Causal-TGAN, leveraging inter-variable causal relationships to improve generated data quality. Zhang et al.~\cite{zhang2021ganblr} offered GANBLR for a deeper understanding of feature importance, and Noock and Guillame-Bert~\cite{nock2022generative} proposed a tree-based approach as an interpretable alternative. Kotelnikov et al.\cite{kotelnikov2022tabddpm} explored tabular data modeling using multinomial diffusion models~\cite{hoogeboom2021argmax} and one-hot encodings, while \textit{FinDiff}~\cite{sattarov2023findiff}, foundational for our framework, uses embeddings for encoding.

\vspace{0.1cm}

\textbf{Federated Deep Generative Models:} De Goede et al. in~\cite{de2023training} devise a federated diffusion model framework utilizing Federated Averaging~\cite{mcmahan2017a} and a UNet~\cite{ronneberger2015u} backbone algorithm to train DDPMs on the Fashion-MNIST and CelebA datasets. This approach reduces the parameter exchange during training without compromising image quality. Concurrently, Jothiraj and Mashhadi in~\cite{jothiraj2023phoenix} introduce \textit{Phoenix}, an unconditional diffusion model that employs a UNet~\cite{ronneberger2015u} backbone to train DDPMs on the CIFAR-10 image database. Both studies underscore the pivotal role of federated learning techniques in advancing the domain. 

\vspace{0.05cm}

To the best of our knowledge, this is the first attempt utilizing diffusion models in the federated learning setup for synthesizing mixed-type tabular financial data.

\section{Federated Learning of Diffusion Models}

This section outlines our proposed \textit{FedTabDiff} model for integrating DDPMs with FL, aiming to enhance the privacy of generated mixed-type tabular data.

\vspace{0.1cm}

\textbf{Gaussian Diffusion Models.} The \textit{Denoising Diffusion Probabilistic Model}~\cite{sohl2015deep, ho2020denoising} is a latent variable model that uses a forward process to perturb data~\scalebox{0.85}{$x_0 \in \mathbb{R}^d$} incrementally with Gaussian noise~\scalebox{0.85}{$\epsilon$}, then restores it through a reverse process. Starting at~\scalebox{0.85}{$x_0$}, latent variables~\scalebox{0.85}{$x_1, \ldots, x_T$} are derived via a Markov Chain, transforming them into Gaussian noise~\scalebox{0.85}{$x_T \sim \mathcal{N}(0, I)$}, defined as:

\vspace{-0.1cm}

\begin{equation}
    q(x_t|x_{t-1})=\mathcal{N}(x_t;\sqrt{1-\beta_t}x_{t-1}, \beta_t I).
    \label{eq:q_step}
\end{equation}

\vspace{0.1cm}

\noindent Here,~\scalebox{0.85}{$\beta_t$} is the noise level at timestep~\scalebox{0.85}{$t$}. Sampling~\scalebox{0.85}{$x_t$} from~\scalebox{0.85}{$x_0$} for any~\scalebox{0.85}{$t$} is expressed as~\scalebox{0.85}{$q(x_t|x_0)=\mathcal{N}(x_t;{\textstyle\sqrt{1-\hat{\beta_t}}} x_0, \hat{\beta_t} I)$}, where~\scalebox{0.85}{$\hat{\beta_t}=1-\prod_{i=0}^{t} (1-\beta_i)$}. In the reverse process, the model denoises~\scalebox{0.85}{$x_t$} to recover~\scalebox{0.85}{$x_0$}. To approximate, a neural network with parameters~\scalebox{0.85}{$\theta$} is trained, parameterizing each step as~\scalebox{0.85}{$p_\theta(x_{t-1}|x_t)=\mathcal{N}(x_{t-1}; \mu_\theta(x_t, t), \Sigma_\theta(x_t,t))$}, where~\scalebox{0.85}{$\mu_\theta$} and~\scalebox{0.85}{$\Sigma_\theta$} are the estimated mean and covariance. Following Ho et al.~\cite{ho2020denoising}, with~\scalebox{0.85}{$\Sigma_\theta$} being diagonal,~\scalebox{0.85}{$\mu_\theta$} is computed as:

\vspace{-0.2cm}

\begin{equation}
    \mu_\theta(x_t, t)=\frac{1}{\sqrt{\alpha_t}}(x_t - \frac{\beta_t}{\sqrt{1-\hat{\alpha_t}}} \epsilon_\theta(x_t, t)).
\end{equation}

\vspace{-0.1cm}

\noindent Here,~\scalebox{0.85}{$\alpha_t := 1-\beta_t$},~\scalebox{0.85}{$\hat{\alpha_t} := \prod_{i=0}^{t} \alpha_i$}, and~\scalebox{0.85}{$\epsilon_\theta(x_t, t)$} is the predicted noise component. Empirical evidence shows that using a simplified MSE loss yields superior results compared to the variational lower bound~\scalebox{0.85}{$\log p_{\theta}(x_0)$}, as given by:

\vspace{-0.2cm}

\begin{equation}
    \mathcal{L}_t=\mathbb{E}_{x_0,\epsilon,t}||\epsilon-\epsilon_\theta(x_t,t)||_2^2.
\end{equation}

\vspace{0.2cm}

\textbf{Federated Learning.} The DDPM learning is enhanced by \textit{Federated Learning}~\cite{mcmahan2017a}. FL enables knowledge acquisition from data distributed across~\scalebox{0.85}{$\mathcal{C}$} individual clients, denoted as~\scalebox{0.85}{$\{\omega_{i}\}^{\mathcal{C}}_{i=1}$}. The training dataset is partitioned into~\scalebox{0.85}{$\mathcal{C}$} sub-datasets,~\scalebox{0.85}{$\mathcal{D}=\{\mathcal{D}_{i}\}^{\mathcal{C}}_{i=1}$}, each privately accessible only to a single client~\scalebox{0.85}{$\omega_{i}$}. Data subsets~\scalebox{0.85}{$\mathcal{D}_{i}$} and~\scalebox{0.85}{$\mathcal{D}_{j}$}, where~\scalebox{0.85}{$i \neq j$}, can have different data distributions. We build on the \textit{FinDiff} model for mixed-type tabular data generation~\cite{sattarov2023findiff}, inspired by the work of de Goede, Jothiraj, and Mashhadi~\cite{de2023training, jothiraj2023phoenix}. We introduce a central \textit{FinDiff} model~\scalebox{0.85}{$f^{\xi}_{\theta}$} with parameters~\scalebox{0.85}{$\theta^{\xi}$}, maintained by a trusted entity and collectively learned by clients~\scalebox{0.85}{$\{\omega_{i}\}^{\mathcal{C}}_{i=1}$}. Each client~\scalebox{0.85}{$\omega_{i}$} maintains a decentralized \textit{FinDiff} model~\scalebox{0.85}{$f^{\omega}_{\theta, i}$} and contributes to learning the central model. The federated optimization uses a synchronous update scheme over~\scalebox{0.85}{$r=1, \dots, \mathcal{R}$} communication rounds. In each round, a subset of clients~\scalebox{0.85}{$\omega_{i, r} \subseteq \{\omega_{i}\}^{\mathcal{C}}_{i=1}$} is selected. These clients receive the current central model parameters~\scalebox{0.85}{$\theta^{\xi}_{r}$}, perform optimization, and then send their updated model parameters back for aggregation. The schematic process with four clients is depicted in  Fig \ref{fig:FedTabDiff}. The \textit{Federated Averaging}~\cite{mcmahan2017a} parameter aggregation technique is used to compute a weighted average over the decentralized model updates, defined as:

\vspace{-0.2cm}

\begin{equation}
    \theta^{\xi}_{r+1} \leftarrow \frac{1}{|\mathcal{D}|} \sum_{i=1}^{\lambda} |\mathcal{D}_{i}| \; \theta^{\;\omega}_{i,r+1} \,,
    \label{equ::federated_learning::federated_averaging}
\end{equation}

\vspace{-0.1cm}

\noindent where~\scalebox{0.85}{$\lambda$} denotes the number of participating clients,~\scalebox{0.85}{$\theta^{\xi}_{r}$} the central and~\scalebox{0.85}{$\theta^{\omega}_{i,r}$} the client model parameters,~\scalebox{0.85}{$r$} the current communication round,~\scalebox{0.85}{$|\mathcal{D}|$} is the total sample count, and~\scalebox{0.85}{$|\mathcal{D}_{i}| \subseteq |\mathcal{D}|$} is the number of samples of client~\scalebox{0.85}{$\omega_i$}. 

\section{Experimental Setup}

This section describes the details of the conducted experiments, encompassing the dataset, data preparation steps, model architecture, and evaluation metrics.

\subsection{Datasets and Data Preparation}
\label{subsec:datasets}

In this preliminary work, the following two real-world tabular datasets are used in our experiments:

\begin{enumerate}
    \item \textbf{Philadelphia City Payments Data}\footnote{\url{https://www.phila.gov/2019-03-29-philadelphias-initial-release-of-city-payments-data}} encompasses a total of 238,894 payments generated by 58 distinct city departments in 2017. Each payment includes 10 categorical and 1 numerical attribute. The feature \textit{doc\_ref\_no\_prefix\_definition} (document reference number prefix definition) was used for the non-iid split.
    \item \textbf{Diabetes Hospital Data}\footnote{\url{https://www.kaggle.com/datasets/brandao/diabetes}} encompasses a total of 101,767 clinical care records collected by 130 US hospitals between the years 1999-2008. Each care record includes 40 categorical and 8 numerical attributes. The feature \textit{age} (a patient's age group) was used for the non-iid split.

\end{enumerate} 

\textbf{Non-iid partition.} We evaluate the proposed \textit{FedTabDiff} model's efficacy and effectiveness, ensuring generalizability across non-iid data splits. The data is non-iid partitioned by segregating based on a categorical feature. The five dominant categories within this feature serve as a basis for data allocation to different clients, thereby introducing a realistic non-iid and unbalanced data environment for federated training. We present the descriptive statistics and details on the non-iid data partitioning scheme in Tab. \ref{tab:datasets}. 

\begin{table}[h!]
    \centering
    \begin{tabular}{@{}c r r@{}}
    \toprule
     & \multicolumn{2}{c}{\textbf{\# samples in $\mathcal{D}_{i}$}} \\
    \cmidrule{2-3}
    \textbf{Client} &  \textbf{Philadelphia} & \textbf{Diabetes} \\
    
    \midrule
     $\omega_1$ & 40,038 & 9,685 \\
     $\omega_2$ & 28,521 & 17,256 \\ 
     $\omega_3$ & 16,831 & 22,483 \\ 
     $\omega_4$ & 93,119 & 26,068 \\ 
     $\omega_5$ & 36,793 & 17,197 \\ 
     \cmidrule{1-3}
     all & 215,302 & 92,689\\ 
    \bottomrule
    \end{tabular}
    \caption{Non-iid and unbalanced data partitioning scheme. Only a subset \scalebox{0.85}{$\mathcal{D}_{i} \subset \mathcal{D}$} is privately accessible to a client \scalebox{0.85}{$\omega_i$}.}
  \label{tab:datasets}
\end{table}

To standardize the numeric attributes, we employed quantile transformations, as implemented in the scikit-learn library.~\footnote{\url{https://scikit-learn.org/stable/modules/generated/sklearn.preprocessing.QuantileTransformer.html}} For the categorical attributes, we utilized embedding techniques following the approach outlined by~\citet*{sattarov2023findiff}.

\subsection{Model Architecture and Hyperparameters}

In the following, we detail the architecture and the specific hyperparameters chosen in \textit{FedTabDiff} model optimization.

\vspace{0.1cm}

\textbf{Model Architecture.} Our architecture for both datasets comprises four layers, each with 1024 neurons. The models are trained for up to \scalebox{0.85}{$R=1,000$} communication rounds utilizing a mini-batch size of 512. We employ the Adam optimizer~\cite{adam} with parameters \scalebox{0.85}{$\beta_{1}=0.9$} and \scalebox{0.85}{$\beta_{2}=0.999$}. Model parameter optimization is conducted using PyTorch v2.0.1~\cite{pytorch}.

\vspace{0.1cm}

\textbf{Model Learning Hyperparameters.} For our \textit{FinDiff} models applied to both datasets, we determined the optimal number of diffusion steps to be \scalebox{0.85}{$T=500$}. A linear learning-rate scheduler is utilized with initial and final rates set at \scalebox{0.85}{$\beta_{start}=0.0001$} and \scalebox{0.85}{$\beta_{end}=0.02$}, respectively. Each categorical attribute corresponds to a 2-dimensional embedding.

\vspace{0.1cm}

\textbf{Federated Learning Hyperparameters.} In each communication round \scalebox{0.85}{$r=1,..., R$}, every client \scalebox{0.85}{$\omega_i$} conducts 20 model optimization updates (client rounds) on its local model \scalebox{0.85}{$\theta^\omega_i$} before sharing the updated parameters. We engage all five clients \scalebox{0.85}{$\lambda=5$} in each communication round for decentralized optimization iteration.\footnote{The federated learning scenario is simulated using the Flower framework v1.4.0~\cite{beutel2022flower}, facilitating multi-client model training.}

\begin{table*}[ht]
  \centering
    \caption{Comparative analysis of the Federated (\textit{FedTabDiff}) versus Non-Federated (\textit{FinDiff}) models, evaluated using the full dataset \scalebox{0.85}{$\mathcal{D}$}. Non-Federated diffusion models are trained individually at each client $\omega_i$ with subset \scalebox{0.85}{$\mathcal{D}_{i} \subset \mathcal{D}$} (reflected in column `Split') and evaluated against the entire dataset~\scalebox{0.85}{$\mathcal{D}$}.}
    \label{tab:quant_results}
    \fontsize{10pt}{10pt}\selectfont    
  \begin{tabular}{@{}l c c c c c c@{}}
    \toprule
    & & & \multicolumn{4}{c}{\textbf{Evaluation Measures}} \\
    \cmidrule{4-7}
    \textbf{Dataset} & \textbf{Client} & \textbf{Split $\mathcal{D}_{i}$}  & \textbf{Fidelity $\uparrow$} & \textbf{Utility $\uparrow$} & \textbf{Coverage $\uparrow$} & \textbf{Privacy $\downarrow$}\\
    \midrule
     \multirow{6}{*}{Philadelphia} 
     & $\omega_1$ & 19\% & 0.267 $\pm$ 0.03 & 0.263 $\pm$ 0.04 & 0.689 $\pm$ 0.03 & 3.162 $\pm$ 0.19 \\
     & $\omega_2$ & 13\% & 0.264 $\pm$ 0.03 & 0.325 $\pm$ 0.06 & 0.681 $\pm$ 0.02 & 3.103 $\pm$ 0.13 \\ 
     & $\omega_3$ & 8\%  & 0.207 $\pm$ 0.03 & 0.118 $\pm$ 0.04 & 0.847 $\pm$ 0.04 & 3.178 $\pm$ 0.03 \\ 
     & $\omega_4$ & 43\% & 0.394 $\pm$ 0.01 & 0.430 $\pm$ 0.01 & 0.863 $\pm$ 0.02 & 2.919 $\pm$ 0.14 \\ 
     & $\omega_5$ & 17\% & 0.238 $\pm$ 0.03 & 0.197 $\pm$ 0.03 & 0.898 $\pm$ 0.01 & 3.359 $\pm$ 0.33 \\ 
     \cmidrule{2-7}
     & \multicolumn{2}{c}{FedTabDiff} & \textbf{0.590} $\pm$ \textbf{0.01} & \textbf{0.837} $\pm$ \textbf{0.03} & \textbf{0.944} $\pm$ \textbf{0.03} & \textbf{2.607} $\pm$ \textbf{0.18} \\ 

    \midrule
     \multirow{6}{*}{Diabetes} 
     & $\omega_1$ & 10\% & 0.217 $\pm$ 0.01 & 0.104 $\pm$ 0.03 & 0.944 $\pm$ 0.02 & 10.261 $\pm$ 0.25 \\
     & $\omega_2$ & 18\% & 0.269 $\pm$ 0.01 & 0.186 $\pm$ 0.01 & 0.943 $\pm$ 0.03 & 10.091 $\pm$ 0.38 \\ 
     & $\omega_3$ & 24\% & 0.314 $\pm$ 0.01 & 0.242 $\pm$ 0.01 & \textbf{0.946} $\pm$ \textbf{0.01} & \ \ 9.895 $\pm$ 0.31 \\ 
     & $\omega_4$ & 28\% & 0.331 $\pm$ 0.01 & \textbf{0.281} $\pm$ \textbf{0.01} & 0.939 $\pm$ 0.01 & \ \ 9.941 $\pm$ 0.21 \\ 
     & $\omega_5$ & 18\% & 0.269 $\pm$ 0.01 & 0.185 $\pm$ 0.01 & 0.943 $\pm$ 0.02 & 10.139 $\pm$ 0.19 \\ 
     \cmidrule{2-7}
     & \multicolumn{2}{c}{FedTabDiff} & \textbf{0.720} $\pm$ \textbf{0.01} & 0.265 $\pm$ 0.01 & 0.906 $\pm$ 0.01 & \textbf{3.120} $\pm$ \textbf{0.09} \\ 

    \bottomrule 
    \multicolumn{7}{l}{\scalebox{0.7}{*Scores are derived from the averaged results and standard deviations of five experiments, each initiated with distinct random seeds}}
  \end{tabular}
\end{table*}

\subsection{Evaluation Metrics}

 We employ a series of standard evaluation metrics, including \textbf{fidelity}, \textbf{utility}, \textbf{coverage}, and \textbf{privacy}, to evaluate the effectiveness of our model quantitatively.\footnote{The implementation of the evaluation metrics originates from the Synthetic Data Vault (SDV) library v1.3.0~\cite{SDV}.} The metrics are selected to represent diverse aspects of data generation quality, ensuring a holistic view of model performance.

\vspace{0.1cm}

\textbf{Fidelity.} Fidelity assesses how closely synthetic data emulates real data, considering both column-level and row-level comparisons. For column fidelity, the similarity between corresponding columns in synthetic and real datasets is evaluated. Numeric attributes employ the \textit{Kolmogorov-Smirnov Statistic} (KSS)~\cite{kolmogorov_1951}, represented as \scalebox{0.85}{$KS(x^d, s^d)$}, to compare empirical distributions. The \textit{Total Variation Distance} (TVD), denoted as \scalebox{0.85}{$\textit{TVD}(x^d, s^d) = \sum_{c\in C} | p(x^{d_c})- p(s^{d_c})|$}, quantifies differences in categorical attributes, with \scalebox{0.85}{$p(s^{d_c})$} indicating the frequency of categories $\mathbf{c}$ in attribute $d$. Column-wise fidelity, \scalebox{0.85}{$\Omega_{col}$}, is then calculated as:

\begin{equation}
    \Omega_{col}=\begin{cases}
        1-KSS(x^d, s^d) & \text{if $d$ is num.} \\
        1-\frac{1}{2}TVD(x^d, s^d) & \text{if $d$ is cat.} 
    \end{cases}
\end{equation}

\noindent The overall fidelity for columns in synthetic dataset $S$ is the mean of \scalebox{0.85}{$\Omega_{col}(x^d, s^d)$} across all attributes. Row fidelity focuses on correlations between column pairs. For numeric attributes, the \textit{Pearson Correlation} (PC)~\cite{benesty2009pearson} between pairs, \scalebox{0.85}{$\rho(x^a, x^b)$}, is used. The discrepancy in correlations for real and synthetic pairs, \scalebox{0.85}{$PC(x^{a,b}, s^{a,b}) = |\rho(x^a, x^b) - \rho(s^a, s^b)|$}, quantifies this aspect. TVD is calculated over category pairs in attributes $a$ and $b$ for categorical attributes.

\begin{equation}
    \Omega_{row}=\begin{cases}
        1-\frac{1}{2}PC(x^{a,b}, s^{a,b}) & \text{if $a$,$b$ are num.} \\
        1-\frac{1}{2} TVD(x^{a,b}, s^{a,b}) & \text{if $a$,$b$ are cat.} 
    \end{cases}
\end{equation}

\noindent The total row fidelity for dataset $S$ is the average of \scalebox{0.85}{$\Omega_{row}(x^{a,b}, s^{a,b})$} across all attribute pairs. Finally, the aggregate fidelity score, formally denoted as \scalebox{0.85}{$\Omega(X, S)$}, is the mean of column and row fidelities.

\vspace{0.1cm}

\noindent \textbf{Utility.} The effectiveness of synthetic data is determined by its utility, a measure of how functionally equivalent it is to real-world data. This utility is quantified by training machine learning models on synthetic datasets and then assessing their performance on original datasets. In this study, we operationalize utility as the performance of classifiers trained on synthetic data ($S_{Train}$) that shares dimensional consistency with the real training set, but then evaluated against the actual test set ($X_{Test}$). This process evaluates the synthetic data's efficacy in replicating the statistical properties necessary for accurate model training. The average accuracy across all classifiers is computed to represent the overall utility of the synthetic data, formalized as:

\vspace{-0.2cm}

\begin{equation}
\Phi = \frac{1}{N} \sum_{i=1}^{N} \Theta_i(S_{Train}, X_{Test}),
\end{equation}


\noindent Here, $\Phi$ represents the utility score, and $\Theta_i$ denotes the accuracy of the $i$-th classifier. To provide a comprehensive evaluation, we selected $N=5$ classifiers for this study, namely Random Forest, Decision Trees, Logistic Regression, Ada Boost, and Naive Bayes.

\vspace{0.1cm}

\noindent \textbf{Coverage.} The coverage metric is essential for evaluating the extent to which a synthetic data column replicates the diversity of categories in a real data column. When applied to categorical features, this metric initially identifies the number of unique categories \scalebox{0.85}{$c_{x^d}$} in the original column. It then assesses whether these categories are represented in the synthetic column \scalebox{0.85}{$c{_s^d}$}. The metric computes the proportion of real categories that are reflected in the synthetic data \scalebox{0.85}{$c_{s^d} \div c_{x^d}$}. For numerical features, where \scalebox{0.85}{$x^d$} and \scalebox{0.85}{$s^d$} denote the real and synthetic columns respectively, the metric evaluates how closely the range of \scalebox{0.85}{$s^d$} (i.e., its minimum and maximum values) aligns with that of \scalebox{0.85}{$x^d$}. This alignment is quantified by:

\vspace{-0.2cm}

\begin{equation}
    \gamma(x^d, s^d)=1 - \left[\max\left(\tau, 0\right) + \max\left(\chi, 0\right)\right],
\end{equation}


\noindent where \scalebox{0.85}{$\tau=\frac{\min(s^d)-\min(x^d)}{\max(x^d)-\min(x^d)}$} and \scalebox{0.85}{$\chi=\frac{\max(x^d)-\max(s^d)}{\max(x^d)-\min(x^d)}$} ensuring a scale-independent measure of $x^d$ and $s^d$.

\vspace{0.1cm}

\noindent \textbf{Privacy}. The privacy metric quantifies the degree to which the synthetic data inhibits the identification of original data entries. The Distance to Closest Records (DCR) is utilized to assess the privacy of the generated data. The DCR is determined as the nearest distance from a synthetic data point \( s_n \) to the authentic data points \( X \), formally expressed as:

\vspace{-0.1cm}

\begin{equation}
    DCR(s_n) = \min_{x\in X} d(s_n, x)
\end{equation}

\vspace{-0.1cm}

\noindent where \( d(\cdot) \) represents the distance metric, with Euclidean distance applied in this context. The final score is the median of the DCRs for all synthetic data points.

\vspace{0.1cm}

\noindent \textbf{Evaluation Process.} The effectiveness of the proposed \textit{FedTabDiff} is assessed by comparing it with a traditional non-federated scenario. In the non-federated scenario, each client independently learns a \textit{FinDiff} model. The quality of the data generated by these models is then assessed in relation to various data partitions and the dataset as a whole. This analysis evaluates the federated model's effectiveness in merging diverse client data insights and enhancing data representation, especially for underrepresented clients with limited data. It assesses the model's ability to produce comprehensive and varied data, which is essential in real-world synthetic data generation.

\section{Experimental results.}

\begin{figure*}[ht]
    \centering
    \begin{subfigure}[b]{0.99\textwidth}
        \centering
        \includegraphics[width=\textwidth]{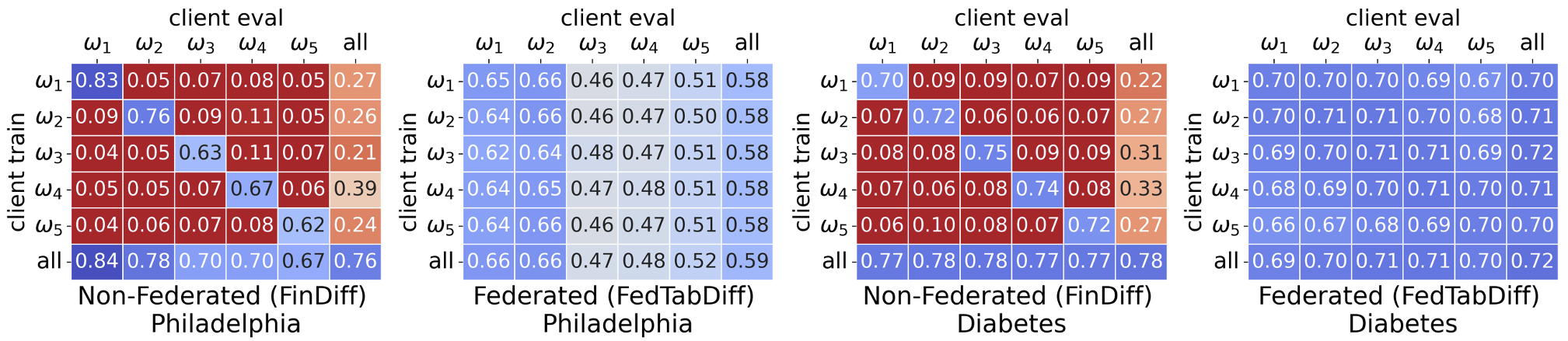}
        \caption{Fidelity scores}
        \label{fig:heatmaps_fidelity}
    \end{subfigure}
    \hfill
    \begin{subfigure}[b]{0.99\textwidth}
        \centering
        \includegraphics[width=\textwidth]{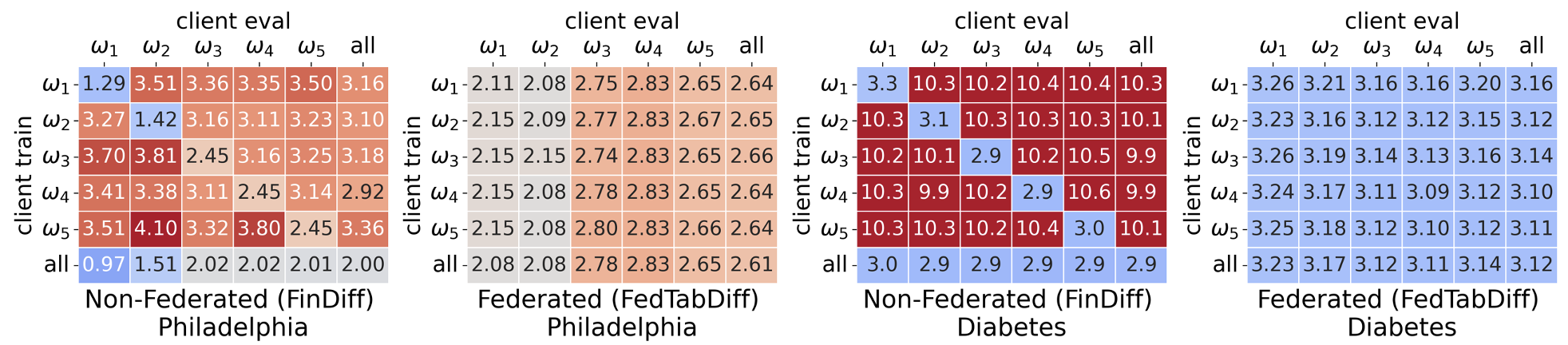}
        \caption{Privacy scores}
        \label{fig:heatmaps_privacy}
    \end{subfigure}
    \hfill
    \begin{subfigure}[b]{0.99\textwidth}
        \centering
        \includegraphics[width=\textwidth]{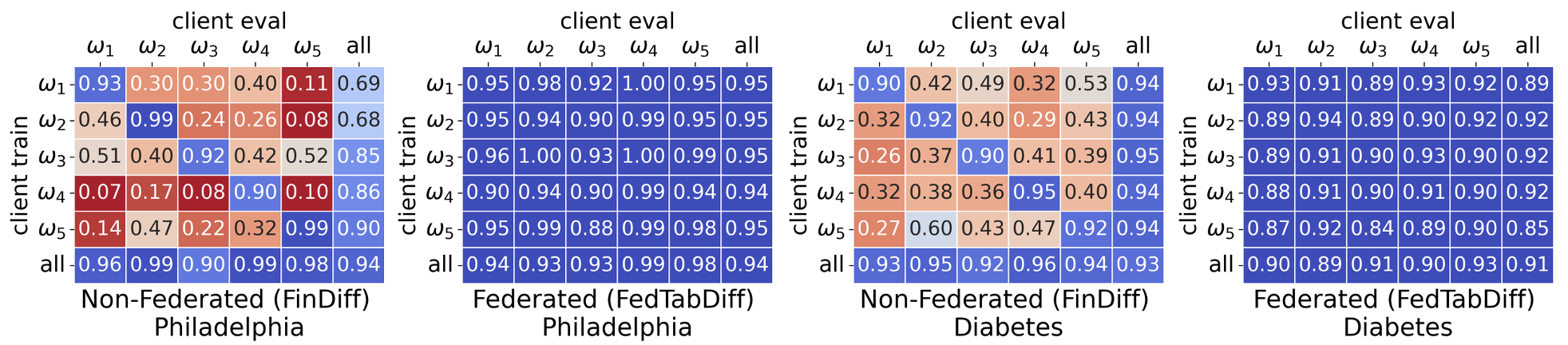}
        \caption{Coverage scores}
        \label{fig:heatmaps_coverage}
    \end{subfigure}
    \caption{Evaluation of Federated (\textit{FedTabDiff}) vs. Non-Federated (\textit{FinDiff}) Diffusion Models in terms of fidelity, privacy, and coverage and across individual clients ($\omega_i$). For the Federated model, the aggregated central model is evaluated across all client data subsets ($\mathcal{D}_1$, $\mathcal{D}_2$, ..., $\mathcal{D}_\lambda$). For the Non-Federated model, each client's model is trained on its respective data subset ($\mathcal{D}_i$) and evaluated across all subsets ($\mathcal{D}_1$, $\mathcal{D}_2$, ..., $\mathcal{D}_\lambda$).}
    \label{fig:heatmaps}
    \vspace{-0.5cm}
\end{figure*}

This section presents the results of the experiments, demonstrating the efficacy of the \textit{FedTabDiff} model and providing quantitative analyses. The results, detailed in \cref{tab:quant_results} together with depicted heatmaps in \cref{fig:heatmaps}, compare the Federated \textit{FedTabDiff} model with Non-Federated \textit{FinDiff} models across fidelity, utility, coverage, and privacy metrics.

\vspace{0.1cm}

\noindent \textbf{Fidelity.} The \textit{FedTabDiff} model exhibits superior fidelity, surpassing non-federated models by up to 56\% for the Philadelphia dataset and 118\% for the Diabetes dataset. Its adaptability is evident across various data subsets, \scalebox{0.85}{$\mathcal{D}_{i}$}, as shown in \cref{fig:heatmaps_fidelity}. \textit{FedTabDiff} consistently achieves stable fidelity scores across all clients $\omega_i$, suitable for the task at hand. In contrast, non-federated models perform comparably only on their training subsets, a limitation visible in the heatmaps' diagonal scores. These models, trained on local subsets \scalebox{0.85}{$\mathcal{D}_i$}, lack exposure to the broader global dataset \scalebox{0.85}{$\mathcal{D}$}.

\vspace{0.1cm}

\noindent \textbf{Utility.} The Utility scores, as detailed in Tab. \ref{tab:quant_results}, show a significant improvement with \textit{FedTabDiff} in the Philadelphia dataset, demonstrating over a 218\% enhancement compared to individual client models. With a score of 0.837, it surpasses all client models, underlining its capability to synthesize data with more comprehensive insights in a federated context. In the Diabetes dataset, \textit{FedTabDiff} is marginally outperformed by client $\omega_4$. This variance can be attributed to the datasets' partitioning characteristics, where the imbalance in the Philadelphia dataset is more pronounced, affecting the federated model's relative performance.

\vspace{0.1cm}
 
\noindent \textbf{Privacy.} The privacy metrics indicate significant improvement with \textit{FedTabDiff}, scoring 2.607 for Philadelphia and 3.120 for Diabetes. While these scores for \textit{FedTabDiff} are promising, it's crucial to scrutinize the chosen privacy metrics to ensure the synthetic data's validity. Generating samples with large deviations from the original data might lead to the production of unrealistic data. Conversely, creating samples too similar to the original may risk data replication, potentially compromising privacy measures. Therefore, a balanced approach, which carefully navigates between these two extremes, is recommended.

\vspace{0.1cm}

\noindent \textbf{Coverage.} For the Diabetes dataset, \textit{FedTabDiff} achieves a score of 0.906, marginally lower than the highest non-federated score of 0.946. Yet, it leads in the Philadelphia dataset with a score of 0.944. These results underscore the performance of \textit{FedTabDiff} and bring nuances in coverage metrics to light. For example, the use of uniform random generation for categorical entries in the Diabetes dataset, as shown in Tab. \ref{tab:quant_results} and the last column of the non-Federated heatmaps in Fig. \ref{fig:heatmaps_coverage}, might lead to inflated scores. On the other hand, while \textit{FedTabDiff} emphasizes the distribution shape, it may inadvertently underrepresent minority entries, particularly in skewed distributions.

\vspace{0.1cm}

In summary, the \textit{FedTabDiff} model consistently outperforms non-federated models across all evaluated metrics, indicating its superior learning capabilities. Harnessing the combined datasets from all clients. leads to a more comprehensive understanding of the underlying data structures, enabling the generation of higher-quality data.

\section{Conclusion}

In this study, we have presented \textit{FedTabDiff}, an innovative federated diffusion-based generative model tailored for the high-quality synthesis of mixed-type tabular data. Its core advantage lies in enabling collaborative model training while upholding the confidentiality of sensitive data.

\vspace{0.1cm}

\textit{FedTabDiff} demonstrates enhanced performance metrics in various datasets, proving its effectiveness in diverse scenarios. Our extensive experimental analysis, conducted in a realistic, non-iid data setting, underscores the methodological robustness of \textit{FedTabDiff}. The model's capability to synthesize tabular data for downstream applications without compromising sensitive information during distributed training is a significant milestone. This aspect is particularly relevant in scenarios where data privacy is paramount.

\vspace{0.1cm}

Looking ahead, future research could explore more sophisticated federated learning frameworks and advanced diffusion models. Such explorations are anticipated to further refine and extend the capabilities of models like \textit{FedTabDiff}, potentially opening new avenues for responsible and privacy-preserving AI applications in various domains.

\section*{Acknowledgements}

Opinions expressed in this work are solely those of the authors and do not necessarily reflect the view of the Deutsche Bundesbank or its staff. M. Schreyer's work is supported by a fellowship within the IFI program (No.: 57515245) of the German Academic Exchange Service (DAAD). 

\bibliography{aaai24}

\end{document}